\documentclass[]{bytedance_seed}

\usepackage[toc,page,header]{appendix}
\usepackage{booktabs}       
\usepackage{threeparttable} 
\usepackage{graphicx}       
\usepackage{caption}         
\usepackage{array}           
\usepackage{amsmath}         
\usepackage{multirow}        
\usepackage{makecell}        
\usepackage{colortbl}        
\usepackage{adjustbox}       
\usepackage{xcolor}          
\usepackage{amssymb}         
\usepackage{amsfonts}        
\usepackage{xspace}
\usepackage{hyperref}

\usepackage{minitoc}
\newcommand{\modelname}{VLA$^3$\xspace}

\title{VLA$^3$: Towards Native Spatial Super-Sensing in \\ Vision Language Action Model }

\author{ByteDance Seed}
\author{Tsinghua University}

\affiliation{Full Author List in \hyperref[sec:author]{Contributions and Acknowledgements}.}

\abstract{
Built upon remarkable advances in large language models (LLM) and vision-language models (VLM), vision-language-action (VLA) models have emerged as a central research paradigm, inheriting extensive visual-language knowledge to enhance robotic generalization. 
However, their reliance on pure 2D data inherently constrains spatial reasoning for the 3D physical world, leading to insufficient spatial understanding that hinders dexterous fine-grained manipulation in complex environments.
Existing solutions--using raw 3D data, auxiliary geometric models, or text-based 3D reasoning data--remain constrained by data scarcity, architectural and feature misalignment, or limited transfer to real-world interaction.
In this work, we present \modelname, a minimally invasive VLA model that achieves native spatial supersensing.
Departing from intricate architectural engineering with dedicated designs, we unlock the inherent geometric potential of a standard mixture-of-transformer VLA backbone via lightweight extensions: an understanding expert for unified multimodal perception and geometric reasoning, and an action expert for continuous robot manipulation. 
The model is jointly optimized under a unified single-stage multi-task paradigm with three complementary objectives: multimodal semantic understanding, metric-space dense geometric reconstruction, and continuous robotic action generation. 
Comprehensive experiments across three core dimensions validate the effectiveness of \modelname: it 
obtains competitive dense reconstruction results compared to specialized geometric models, and delivers notable improvements in real-robot manipulation success rates over VLA baselines, while retaining strong general multimodal reasoning capabilities.
}

\date{\today}

\begin{document}
\maketitle


\section{Introduction}


The core objective of Embodied AI \cite{gupta2021embodied} (also known as Physical AI) is to develop an embodied multimodal model for robots, capable of perception, interaction, and planning with the physical environment.
Inspired by the substantive progress of Large Language Models (LLMs \cite{singh2025openai,guo2025deepseek,seed2026seed1,seed2026seed2,yang2025qwen3}) and Vision-Language Models (VLMs \cite{achiam2023gpt,lu2024deepseek,guo2025seed1,bai2025qwen3}), Vision-Language-Action models (VLAs \cite{brohan2023rt,kim2024openvla,black2024pi_0}) have recently emerged as a central research focus in the community.
These VLA models aim to inherit the extensive visual-language knowledge from VLMs for enhancing the generalization of robotic strategies \cite{zhang2026vlm4vla,fang2025robix}.
The majority of existing models have focused on improving the spatial understanding ability \cite{fang2025robix}, developing more advanced network architectures \cite{li2024vision,zhou2025vision,shi2026memoryvla,hu2026bagelvla}, incorporating additional modalities \cite{zhen20243d,zhang20264d,wang2026lamp,zhong2026flowvla}, and refining action decoding experts \cite{pertsch2025fast,wen2025dexvla}.
Despite these gains, the exclusive dependence on 2D visual data constrains spatial reasoning in the broader realm of the 3D physical world.
In consequence, current works still struggle with the core spatial skills that require geometric understanding \cite{fan2026vlm}, even for simple tasks like spatial distance estimation.
These capabilities are crucial for robots to operate in complicated scenes, such as
daily environments \cite{liu2024visual,uppal2024spin,yang2024equibot}, and to perform dexterous and fine-grained manipulation skills \cite{qin2023dexpoint,wang2024dexcap}.

\begin{figure}[t]
  \centering
  \includegraphics[width=\textwidth]{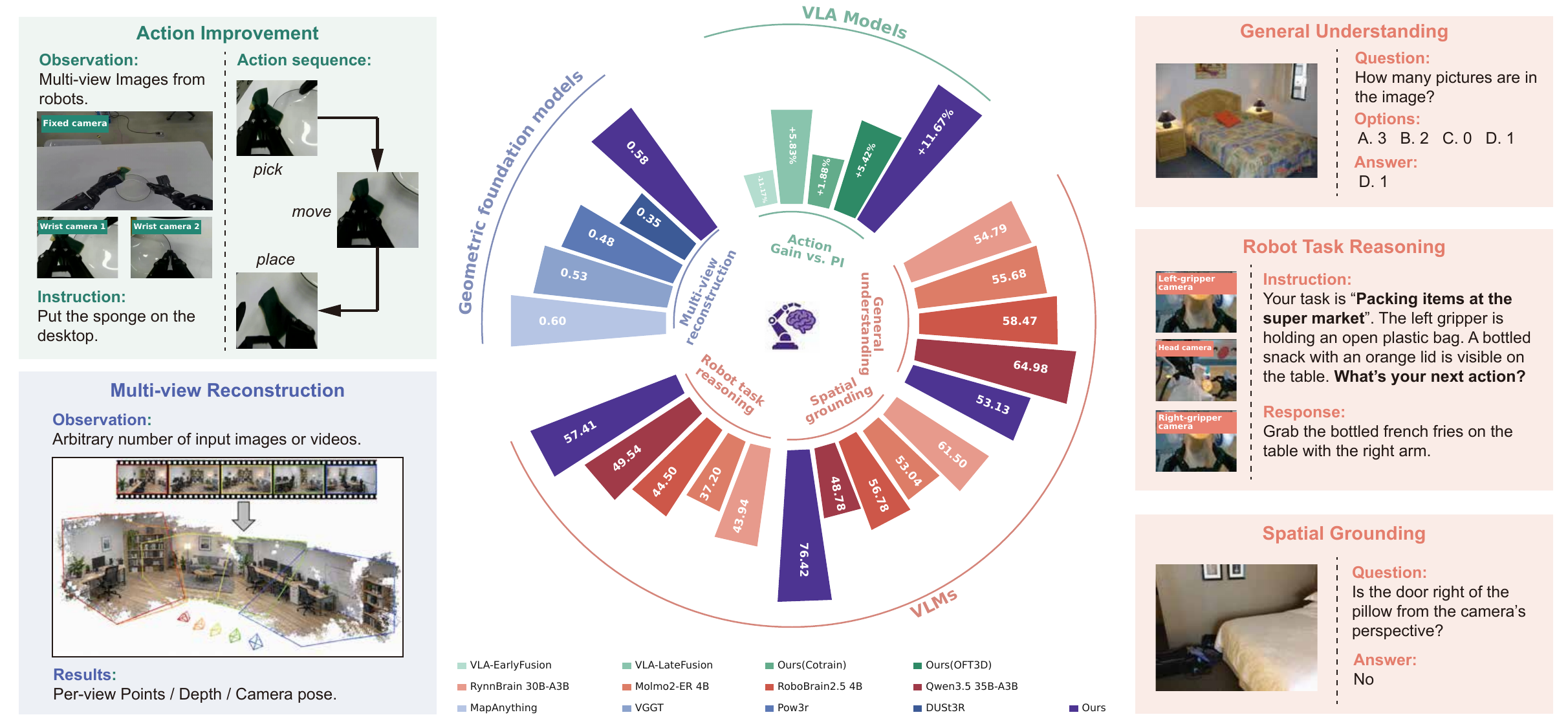}
  \caption{Our VLA model jointly supports three tasks: vision-question answering (VQA), multi-view geometric reconstruction, and action-policy prediction. On VQA benchmarks, it achieves state-of-the-art performance on spatial and robot-related reasoning tasks while maintaining competitive general multimodal understanding. For multi-view reconstruction, it performs comparably to or better than specialized geometric foundation models, as measured by the mean of $1-AbsRel$ across multiple datasets. On robotic manipulation tasks, it improves the success rate by 11.67\% over the baseline model PI.}
  \label{fig:intro}
\end{figure}


In pursuit of human-like visual spatial intelligence, the research community has explored three major paradigms to extend VLM/VLA with 3D spatial understanding.
(1) \textit{Explicit infusion of additional 3D data}: Structured 3D representations (e.g., point clouds) are directly fed into VLM/VLA models \cite{han2026stereopolicy,li2026pointvla,deng2025stereovla,fu2024scene,hong20233d,zheng2025video}, requiring dedicated 3D encoders or cross-modal fusion modules.
Despite the great 3D data efficiency \cite{han2026stereopolicy,ze20243d} compared with 2D data, the scarcity of high-quality 3D data still severely constrains the representational capacity of these models.
(2) \textit{Implicit integration of additional geometric experts}: Spatial features from geometric foundation models (VGGT \cite{wang2025vggt}, Depth-Anything-3 \cite{lin2025depth}, etc.) are fused into VLM/VLA backbones via various integration schemes, including early-fusion \cite{lin2025evo, abouzeid2025geoaware, zheng2026learning, yang2026understanding, yu2026_3dmix}, late-fusion \cite{zhang2025spatial, yang2026abot, yang2026understanding, yu2026_3dmix}, mixture-of-transformers \cite{wang2026lamp}, and etc.
Despite the measurable improvements in spatial reasoning, these efforts inevitably impose invasive modifications on the pretrained models, inducing extra feature alignment issues and extra computation overhead.
(3) \textit{Offline construction of 3D reasoning data}: 3D spatial priors are converted into text QA pairs for VLM fine-tuning \cite{cai2026depthlm,cai2026vlm3}.
While these approaches enable non-invasive injection of spatial priors into VLMs, such text-based data does not constitute authentic geometric features.
It mainly boosts QA performance but fails to support spatially grounded physical reasoning, leading to marginal gains when adapted to interactive VLA models (Table 5).

In this work, we step back from established spatial understanding regimes in embodied foundation models and return to a more fundamental goal:
\textbf{Forsaking elaborate architectural engineering, can we find a minimal modeling strategy to simultaneously integrate spatial understanding with VLA models?}
Our work provides an affirmative answer to this question. 
The core insight is that \textit{a single plain VLA model suffice for native visual geometry grounded understanding, without excessive incorporation of a geometric expert with 3D inductive bias}.
The key lies in transforming VLA models into native geometric foundation models via minimal extension by appending geometric decoders to the backbone, and co-training the model with dense reconstruction objectives, text-based QA objectives, and action learning objectives.
We demonstrate that this simple and minimal regime is sufficient to reconstruct 3D space like a geometric foundation model, advance spatial understanding and robotic manipulation in VLA models, as shown in Figure \ref{fig:intro}. 

We present \textbf{\modelname}, a native \textbf{S}patial \textbf{S}uper-\textbf{S}ensing \cite{yang2026cambrian} \textbf{VLA} model with a minimal architectural design.
\modelname unlocks the inherent geometric potential within a standard Mixture-of-Transformer (MoT) VLA backbone through only lightweight extensions, maximally preserving the rich visual-language priors from large-scale 2D pre-training.
Our model is comprised of two experts: an understanding expert for multimodal perception and geometric reasoning, and an action expert for continuous robot manipulation.
The understanding expert is a 40-layer Mixture-of-Experts (MoE) transformer, which processes visual tokens, text tokens, and learnable geometric tokens (camera/scale) in a unified sequence.
The native visual tokens predicted by the understanding expert are fed to a lightweight DPT \cite{ranftl2021vision} decoder to predict dense depth maps, while the learnable geometric tokens are fed to two MLP heads to regress camera poses and metric scale factors.
In this way, the understanding expert with a backbone of VLM can yield geometrically consistent metric spatial reconstructions without any geometric foundation experts.
In parallel, the action expert learns to generate continuous manipulation trajectories via a flow matching objective.
The training corpus comprises three categories of data: (1) large-scale visual question-answering pairs covering general, spatial, and robot-relevant scenarios; (2) multi-view image data with depth and camera pose annotations in real-world metric; (3) real-world robot demonstration data with state-action trajectories.
We adopt a stratified sampling strategy to interleave these data streams, balancing the learning pace and gradient contributions across different tasks.

We conduct comprehensive evaluation experiments to validate \modelname across three core dimensions: dense geometric reconstruction, real-world robot manipulation, and vision-language spatial understanding.
The experimental results on these benchmarks demonstrate that \modelname obtains competitive reconstruction performance with expertise models under strict OOD settings, yields notable improvements in manipulation success rates against VLA baselines, and achieves state-of-the-art performance on spatial understanding benchmarks, while retaining strong general multimodal reasoning capabilities compared to existing VLMs.
The primary contributions of this work are summarized as follows:
\begin{itemize}
    \item We propose \modelname, a multi-modal VLA model that achieves native spatial supersensing via a minimally invasive architectural design.
    \item We reveal that a single plain VLA model is inherently capable of geometrically grounded 3D understanding, without dedicated external geometric foundation models.
    \item \modelname obtains competitive dense reconstruction results under strict out-of-distribution evaluation, yields consistent improvements in real-robot manipulation tasks, and achieves state-of-the-art performance on multiple spatial understanding benchmarks, demonstrating effective transfer of spatial geometric knowledge to physical interaction.
\end{itemize}




\section{Approach}
Our goal is to endow VLA models with 3D geometric knowledge, thereby enhancing spatial understanding, robotic task-level reasoning, and action planning. To this end, \modelname is trained with three complementary learning objectives: 1) It is trained on visual question-answering data aggregated from diverse sources to strengthen robot-relevant scene understanding and reasoning. 2) It learns dense multi-view geometric reconstruction, enabling the acquisition of robust 3D representations and spatial perception capabilities. 3) It leverages real-world robotic interaction data to translate perceptual and reasoning capabilities into effective manipulation policies. 

\begin{figure}[t]
  \centering
  \includegraphics[width=\textwidth]{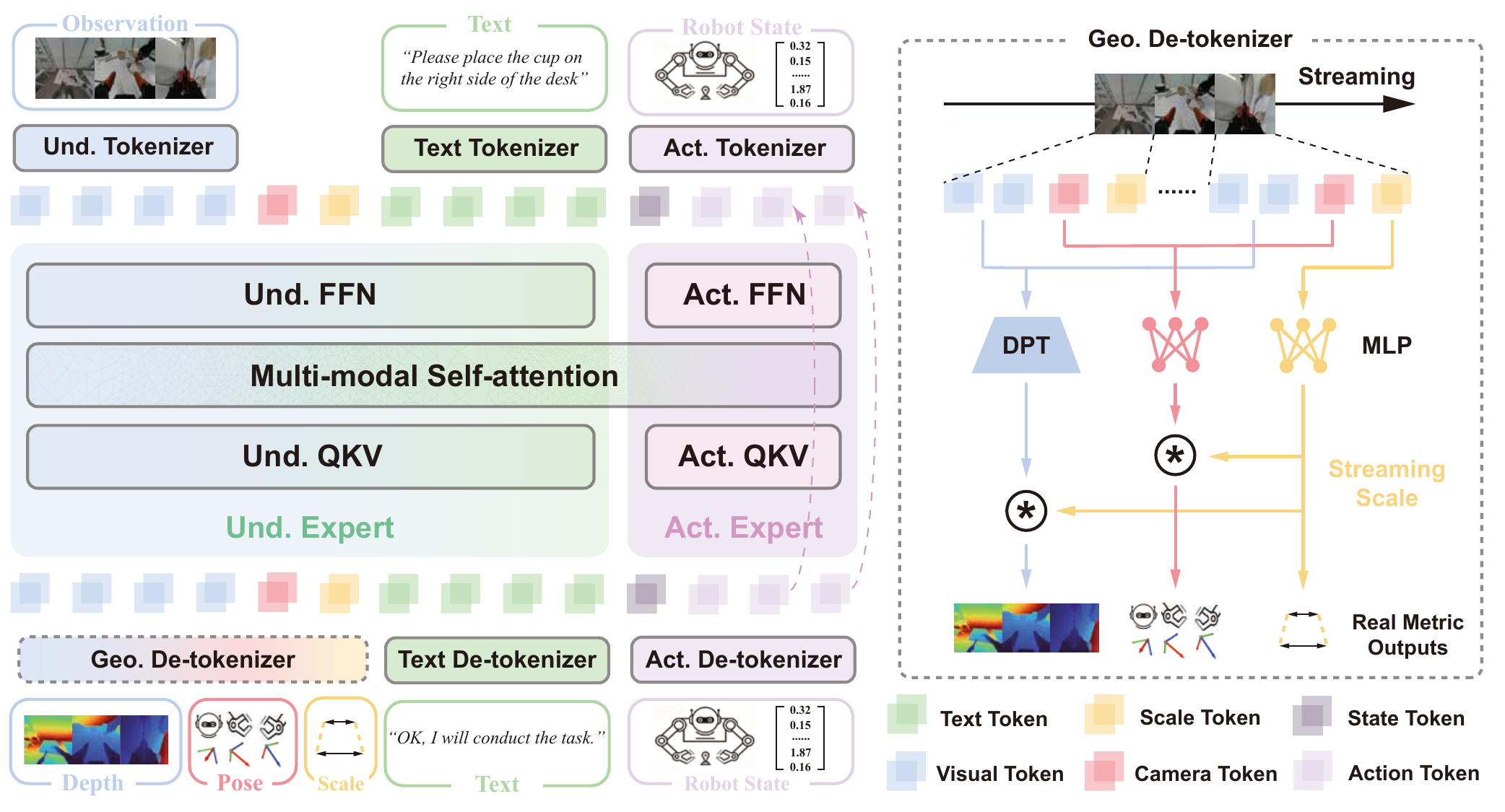}
  \caption{Illustration of \modelname.}
  \label{fig:method}
\end{figure}

\subsection{\modelname}
As illustrated in Figure \ref{fig:method}, \modelname adopts a Mixture-of-Transformers (MoT) architecture comprising two specialized components: an understanding expert and an action expert \cite{black2410pi0,Lv2025F1AV,shi2024lmfusion}. Textual and visual inputs are first converted into token representations and processed by the understanding expert. In addition, a set of learnable camera and scale tokens is introduced to support 3D multi-view reconstruction, including metric depth and camera-pose estimation. In parallel, the robot’s historical states and noise-perturbed future continuous actions are encoded as state and action tokens, respectively, and processed by the action expert. Formally, conditioned on visual tokens $\bm{v}$, text/instruction tokens $\bm{t}$, learnable camera tokens $\bm{c}$, learnable scale tokens $\bm{\sigma}$, robot state tokens $\bm{s}$, and noised action tokens $\bm{a}_\tau$, \modelname models the joint distribution of textual outputs $\bm{o}$, per-view metric depths $\bm{d}$, per-view metric poses $\bm{p}$, and future robot action chunk $\bm{a}$:
\begin{equation}
P_{\theta}\!\left(
\bm{o}, \bm{d},\bm{p}, \bm{a}
\mid
\bm{v},\bm{t},\bm{c},\bm{\sigma},\bm{s}, \bm{a}_\tau
\right).
\end{equation}

Both the understanding and action experts consist of 40 Transformer layers. The understanding expert adopts a Mixture-of-Experts (MoE) \cite{shazeer2017outrageously} architecture to provide high model capacity while maintaining computational efficiency. It contains 34B parameters in total, of which approximately 2.5B are activated for each token during a forward pass \cite{guo2025seed1}. By contrast, the action expert is implemented as a lightweight dense counterpart of the understanding expert. It preserves the same 40-layer depth but removes the MoE modules, resulting in 2B total—and fully activated—parameters.

The heterogeneous tokens are passed through several MoT layers to enable layer-wise interaction, where each layer can be formalized as:
\begin{equation}
\bm{h}_{\bm{v}}, \bm{h}_{\bm{t}}, \bm{h}_{\bm{c}}, \bm{h}_{\bm{\sigma}}, \bm{h}_{\bm{s}}, \bm{h}_{\bm{a}_\tau} = \mathrm{MHSA} ([\mathrm{QKV}_{und}(\bm{v},\bm{t},\bm{c},\bm{\sigma}), \mathrm{QKV}_{act}(\bm{s}, \bm{a}_\tau)]).
\end{equation}
Here, $\bm{h}$ denotes the hidden states, with the subscripts indicating the corresponding token modalities. The expert-specific QKV projections map the hidden representations of the understanding and action streams into queries, keys, and values, which are subsequently processed by multi-head self-attention (MHSA) \cite{vaswani2017attention}. Although the two experts employ separate projection and feed-forward layers, their hidden representations are concatenated into a unified multimodal sequence for global self-attention. Specifically, MHSA uses a causal mask, allowing each token to attend only to itself and preceding tokens while facilitating information exchange across modalities and expert streams.

\textbf{Understanding}.
The hidden states $\bm{h}_{\bm{t}}$ are projected onto the vocabulary space through an LM head, yielding a categorical distribution over candidate tokens. A next-token cross-entropy objective is defined, where the loss is computed only over the target response tokens:
\begin{align}
P({\bm{o}}) = & \mathrm{LMHead}(\bm{h}_{\bm{t}}), \\
\mathcal{L}_{und} = & \mathbb{E}_{\bm{o}_i} [-\mathrm{log}(P({\bm{o}}_i | \bm{o}_{<i}))].
\end{align}

\textbf{Multi-view reconstruction}.
The metric depths and poses are decoded from visual hidden states $\bm{h}_{\bm{v}}$, camera hidden states $\bm{h}_{\bm{c}}$, and scale hidden states \(\bm{h}_{\bm{\sigma}}\). Specifically, a DPT-based decoder \cite{ranftl2021vision} aggregates multi-level visual features extracted from the $10^{th}$, $20^{th}$, $30^{th}$, and $40^{th}$ Transformer layers to predict scale-ambiguous depth maps. In parallel, two MLP-based prediction heads decode the camera poses and scale factors from $\bm{h}_{\bm{c}}$ and $\bm{h}_{\bm{\sigma}}$, respectively.
Our model employs causal attention, whereas conventional geometric foundation models typically use bidirectional global attention \cite{wang2025vggt}. Consequently, each frame attends only to itself and preceding views. As the observed scene context expands, this causal formulation may lead to scale inconsistencies across frames. 
To mitigate this issue, the predicted per-frame scale factors are subsequently applied to the depth maps and the translational components of the camera poses, yielding geometrically consistent outputs in the first frame’s camera coordinate system at real-world metric scale:
\begin{align}
{\bm{\sigma}} = & \mathrm{MLP}(\bm{h}_{\bm{\sigma}}), \\
{\bm{p}} = & \mathrm{MLP}(\bm{h}_{\bm{c}}) * {\bm{\sigma}}, \\
{\bm{d}} = & \mathrm{DPT}(\bm{h}_{\bm{v}}(i) \mid_{i=10,20,30,40}) * {\bm{\sigma}}
\end{align}
Given ground truth depths $\bm{d}_{\rm{gt}}$ and poses $\bm{p}_{\rm{gt}}$, the loss is defined as:
\begin{align}
\mathcal{L}_{geo} = \mathbb{E}_{\bm{p},\bm{d}} [||\bm{p}_{\rm{gt}}  - \bm{p}||_1+||\bm{d}_{\rm{gt}}  - \bm{d}||_1].
\end{align}

\textbf{Action planning}.
The action chunk $\bm{a}$ is modeled by a flow matching process \cite{lipman2022flow}, where the action expert learns to reverse a gradual noise-addition process added in the forward process. During training, a random noise $\epsilon \sim \mathcal{N}(\bf{0}, \bf{I})$ and a timestep $\tau \in [0, 1]$ are sampled to model the noised actions:
\begin{equation}
\bm{a}_{\tau} = \tau \bm{a} + (1 - \tau) \epsilon.
\end{equation}
The denoising velocity vector $\bm{u}_{\tau}$ is decoded by an MLP from the last action hidden states $\bm{h}_{\bm{t}}$. The model is optimized by minimizing the predicted and target velocity vectors: 
\begin{align}
\bm{u}_{\tau} = & \mathrm{MLP}(\bm{h}_{\bm{a}_{\tau}}), \\
\mathcal{L}_{act} = & \mathbb{E}_{\bm{u}_{\tau}^{}}[|| \bm{u}_{\tau}^{} - (\epsilon - \bm{a}) ||_2].
\end{align}

\subsection{Training recipe}
\modelname is trained in a unified single-stage multi-task framework, with its pretrained parameters initialized from the VLM model Robix \cite{fang2025robix}. During training, three complementary objectives—multimodal understanding, multi-view reconstruction, and robotic action generation—are jointly optimized. The loss is formalized as:
\begin{equation}
\mathcal{L}_{{total}} = \alpha \cdot \mathcal{L}_{{und}} + \beta \cdot \mathcal{L}_{{geo}} + \mathcal{L}_{{act}},
\end{equation}
where $\alpha=0.01$, $\beta=0.1$. Accordingly, the training corpus comprises three categories: (1) visual question-answering pairs for general, spatial, and robot-relevant understanding, (2) multi-view geometric data for depth and camera-pose reconstruction, 
and (3) robot demonstration data for action learning. We interleave these data sources using an understanding-to-reconstruction-to-action sampling ratio of 0.5:0.5:1, corresponding to normalized sampling probabilities of 25\%, 25\%, and 50\%, respectively.

\subsubsection{Data engine}
We collect large-scale visual question–answer data from the internet and multi-view geometric data from publicly available repositories. In addition to these sources, we construct a real-world dataset using our own robotic platform, pairing 3D geometric annotations, including metric depth and camera poses, with robot demonstrations comprising state observations and continuous action trajectories.
Our robot, equipped with two arms, integrates a head-mounted camera, two hand-mounted cameras associated with the left and right grippers, and inertial measurement unit (IMU) sensing.

\textbf{3D data}. We apply a tailored processing pipeline to label accurate and dense geometry for real-robot data.
Firstly, VGGT~\cite{wang2025vggt} is adopted to generate per-frame camera poses and depth maps.
Secondly, MegaSAM~\cite{li2025megasam} is applied to jointly refine the predicted depth and camera poses as a post-processing step.
For each sequence, we use the head camera in the first frame and its depth map as the reference camera and reference depth, respectively. We express all views in the reference camera’s coordinate system, establishing a shared spatial reference that accounts for camera motion during data curation. To recover a consistent metric scale, we align the predicted depths and camera poses with the known multi-view extrinsics of the head cameras and the coarse pose estimates.
Finally, we reduce flying points by thresholding the per-pixel confidence maps
predicted by MapAnything~\cite{keetha2026mapanything}, discarding depth whose confidence
falls in the lowest $25\%$ of each frame.

\textbf{Robot data}.
Robot demonstrations are collected through tele-operation, allowing a human operator to guide the platform through manipulation tasks \cite{mandlekar2019scaling,zhao2023learning}. Action data are derived from a pose-estimation pipeline that incorporates IMU measurements and the robot’s kinematic description encoded in the Unified Robot Description Format (URDF).

\subsubsection{Understanding}
\textbf{Spatial understanding}.
We construct visual question–answer pairs from geometric labels, covering multiple task types:
(1) 3D bounding box detection: predicting 3D bounding boxes from monocular images with open-vocabulary object descriptions;
(2) 4D bounding box detection: the same object-level formulation on video, with 3D boxes expressed in the first-frame camera coordinate system;
(3) object-camera motion prediction: inferring object motion relative to the camera;
(4) relative depth sorting: inferring the depth ordering of query pixels within a single image;
(5) absolute depth estimation: estimating metric depth at query pixels, with a single query, two queries, or a query conditioned on a reference pixel of known depth;
(6) ego-motion prediction: modeling camera motion over time to support temporal and spatial reasoning. 
Training samples are constructed from public 3D datasets--including TartanAir \cite{wang2020tartanair}, Waymo \cite{sun2020scalability}, ScanNet \cite{dai2017scannet}, Parallel Domain 4D \cite{van2024generative}, Kubric 4D \cite{van2024generative}, Hypersim \cite{roberts2021hypersim}, DL3DV \cite{ling2024dl3dv}, CO3D \cite{reizenstein2021common}, BlendedMVS \cite{yao2020blendedmvs}, Aria Synthetic Environments \cite{pan2023aria}, and Aria Digital Twins \cite{pan2023aria}--as well as simulated and real-robot data with reconstruction pseudo-labels generated by the 3D data engine. 

\textbf{General Multimodal Understanding and Reasoning}. 
To preserve broad vision-language competence during embodied training, we incorporate several general-purpose multimodal data sources covering image and video captioning, visual question answering, factual and domain-specific knowledge, multi-image interleaved understanding, and instruction following. The corpus includes bilingual image-text data, encyclopedic and scientific knowledge, medical imagery, and document-centric tasks involving OCR, tables, charts, diagrams, and graphical user interfaces. 
Together, these sources provide a general multimodal foundation that supports robust recognition, language grounding, and cross-domain generalization. 

\textbf{2D Visual Grounding}. 
We curate several data sources to strengthen the model’s ability to associate language instructions with spatially localized visual entities. 
We train on four complementary formats: (1) point annotations, where the model marks object centers or part locations from open-vocabulary descriptions, or identifies the referred region given coordinates; (2) 2D bounding-box annotations, including both language-to-box localization and box-to-language region description, sourced from collections such as Visual Genome \cite{krishna2017visualgenome}, Objects365 \cite{shao2019objects365}, OpenImages \cite{kuznetsova2020openimages}, COCO \cite{lin2014coco}, RefCOCO \cite{kazemzadeh2014referitgame,yu2016modeling}, and internal Seed-1.5-VL grounding corpora \cite{guo2025seed15vl}; (3) counting, which requires localizing all instances of a category (often via a two-stage mark-then-count pipeline) before reporting the total; and (4) visual prompts, in which points, boxes, or other overlays are fused with text so the model can reason about marked regions. 

\textbf{Robot Action Understanding}. 
We include robot-centric VLM data sources that connect visual observations and natural-language instructions to high-level action semantics. 
The mix covers: (1) instruction-conditioned VLA-style prediction, in which the model outputs discretized end-effector deltas and gripper states from egocentric images and a language goal, using public manipulation corpora such as Bridge \cite{walke2023bridgedata}, Fractal \cite{brohan2023rt1}, DROID \cite{khazatsky2024droid}, LIBERO \cite{liu2023libero}, RoboCasa \cite{nasiriany2024robocasa}, and AgiBot \cite{agibot2025world}, as well as internal dual-arm packs; (2) meta-action prediction, which abstracts short-horizon motion into compact directional/gripper codes (e.g., left/right arm translation and gripper state over the next 1–2 seconds) from multi-camera observations; (3) chain-of-thought action traces that first reason about the scene and then emit structured actions, including multi-turn variants. 

\subsubsection{Multi-view reconstruction}
Following the geometric foundation models like VGGT \cite{wang2025vggt} and Depth-Anything-3 \cite{lin2025depth}, we curate a collection of 3D/4D datasets to scale up the training corpus, including: Spring \cite{mehl2023spring}, Unreal4K \cite{tosi2021smd}, WildRGBD \cite{xia2024rgbd}, Waymo \cite{sun2020scalability}, TartanAir \cite{wang2020tartanair}, ScanNet \cite{dai2017scannet}, vKITTI2 \cite{cabon2020virtual}, SAIL-VOS 3D \cite{hu2021sail}, ScanNet++ \cite{yeshwanth2023scannet++}, PointOdyssey \cite{zheng2023pointodyssey}, Infinigen Indoors/Nature \cite{raistrick2024infinigen}, InsScene \cite{li2025iggt}, MPSD \cite{antequera2020mapillary}, ParallelDomain4D \cite{van2024generative}, MapFree \cite{arnold2022map}, MegaDepth \cite{li2018megadepth}, Kubric4D \cite{van2024generative}, MVS-Synth \cite{huang2018deepmvs}, ETH3D \cite{schops2017multi}, Hypersim \cite{roberts2021hypersim}, DL3DV \cite{ling2024dl3dv}, CO3D v2 \cite{reizenstein2021common}, BEDLAM \cite{black2023bedlam}, Dynamic Replica \cite{karaev2023dynamicstereo}, DTU \cite{aanaes2016large}, 7-Scenes Training \cite{brahmbhatt2018geometry}, BlendedMVS \cite{yao2020blendedmvs}, Aria Synthetic Environments \cite{pan2023aria}, Aria Digital Twins \cite{pan2023aria}, and an internal dataset of simulated and real-robot data with reconstruction pseudo-labels generated by the 3D data engine.
The collection spans indoor, outdoor, object-centric, robotics, and diverse in-the-wild scenes.
Similar to VGGT \cite{wang2025vggt}, we apply uniform sampling over the aforementioned datasets to construct training samples and maintain balanced coverage over diverse data distributions.
The combination of these datasets is broadly comparable to those of MASt3R \cite{duisterhof2025mast3r} and VGGT \cite{wang2025vggt} in size and diversity.



\subsubsection{Action planning}
We train the low-level control policy on large-scale tele-operated demonstrations from a dual-arm mobile manipulator.
Each sample consists of a language instruction, three calibrated RGB views (head-mounted and left/right wrist cameras), a short-horizon action chunk and the robot state.
The corpus has three complementary sources: (1) standard tele-operation: tabletop and kitchen skills such as sorting tableware, storing toys and blocks, placing fruit, clearing trash, and rearranging bottles; (2) long-horizon task packs: multi-stage household routines (e.g., tableware, stationery, and bedding packs) that require sequencing several sub-skills. (3) human-takeover traces: interventions collected when an autonomous policy fails, emphasizing recovery from distribution shift.

\begin{figure}[t]
  \centering
  \includegraphics[width=\textwidth]{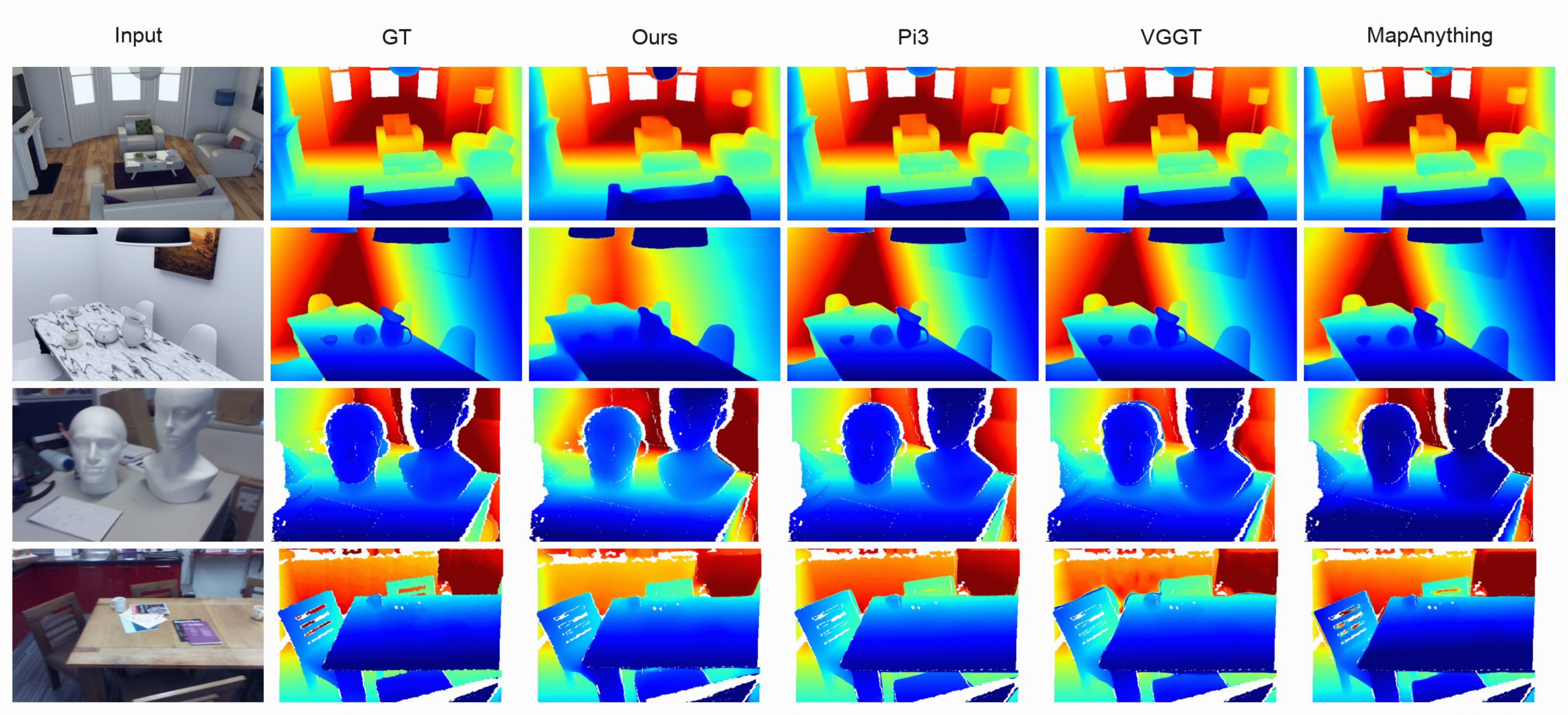}
  \caption{Visualization of depths.}
  \label{fig:depth}
\end{figure}

\section{Experiments}
We conduct extensive experiments to comprehensively evaluate \modelname across three key dimensions: dense geometric reconstruction, robotic manipulation, and vision–language understanding. We focus on the following key questions:
\begin{itemize}
    \item Can \modelname reconstruct dense 3D geometry from multi-view images?
    \item Do these geometric and spatial capabilities translate into improved performance on robotic manipulation tasks?
    \item What is the most effective way to incorporate 3D geometric capabilities into VLA models?
    \item Does learning dense 3D reconstruction improve 3D spatial understanding in vision–language tasks?
    \item How does incorporating geometric supervision for 3D reconstruction affect general vision–language understanding?
\end{itemize}

\begin{figure}[t]
  \centering
  \includegraphics[width=\textwidth]{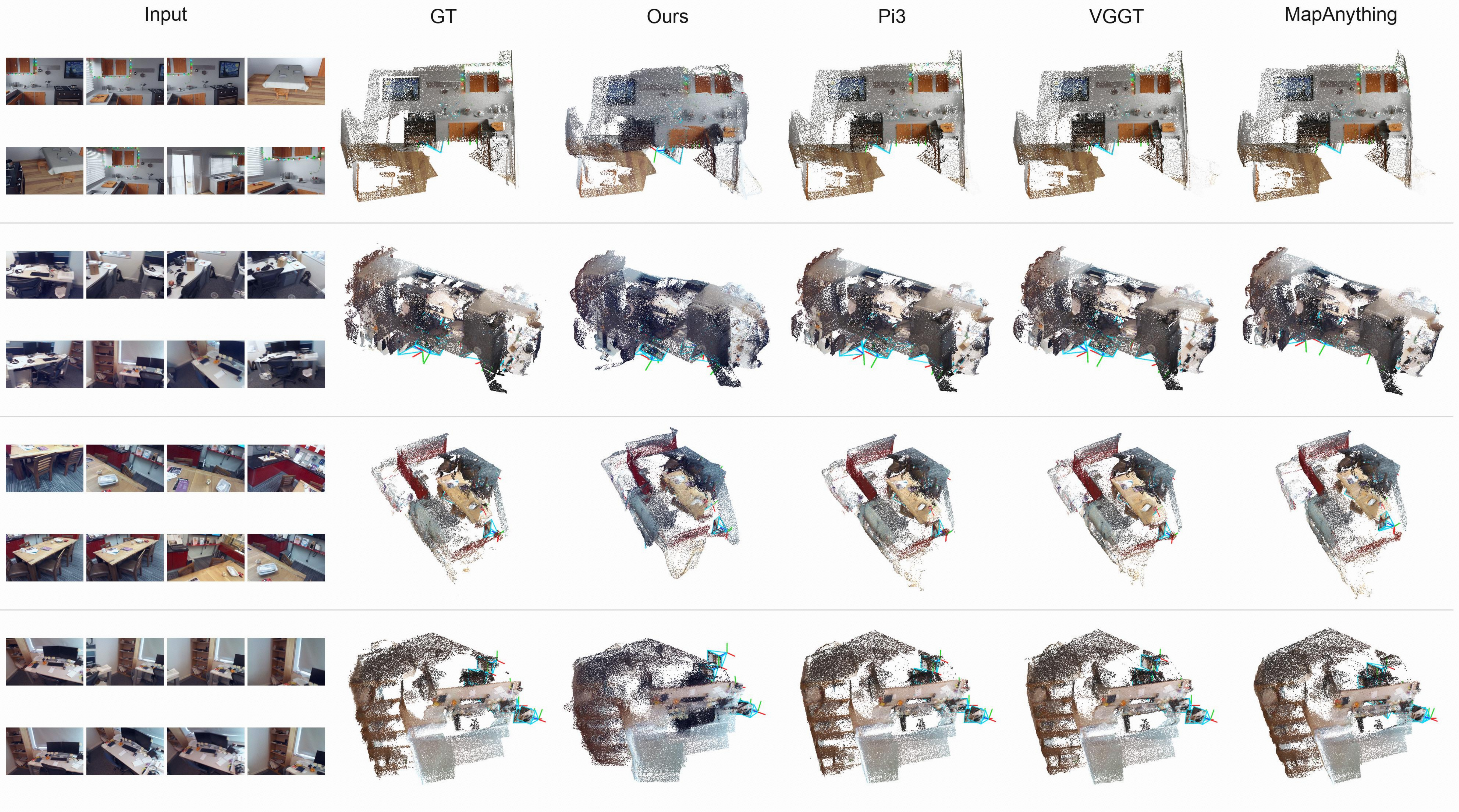}
  \caption{Visualization of point clouds.}
  \label{fig:point}
\end{figure}

\subsection{Multi-view reconstruction}

We evaluate \modelname on dense depth prediction and camera pose estimation across five datasets to assess its ability to reconstruct dense 3D geometry from multi-view images, with results reported in Tables \ref{tab:pose} and \ref{tab:depth}.
We compare \modelname with several geometric foundation models, including DUSt3R \cite{wang2024dust3r}, VGGT \cite{wang2025vggt}, Pi3 \cite{wang2026pi}, Pow3r \cite{jang2025pow3r}, VGGT-Omega \cite{wang2026vggt}, Depth-Anything-3 \cite{lin2025depth}, and MapAnything \cite{keetha2026mapanything}. All results use two input views.
We evaluate on one in-domain dataset (ScanNet++ \cite{yeshwanth2023scannet++}), and four out-of-distribution (OOD) datasets (Neural RGB-D \cite{azinovic2022neural}, 7Scenes \cite{shotton2013scene}, Sintel \cite{butler2012naturalistic}, and TUM \cite{sturm2012benchmark}) to assess both in-domain performance and cross-dataset generalization.

\begin{table}[t]
\centering
\setlength{\tabcolsep}{3.5pt}
\renewcommand{\arraystretch}{1.12}
\resizebox{1\textwidth}{!}{%
\begin{tabular}{l cc cc cc cc cc}
\toprule
\multirow[t]{2}{*}{\textbf{Model}}
  & \multicolumn{2}{c}{\textbf{ScanNet++}}
  & \multicolumn{2}{c}{\textbf{Neural RGB-D}}
  & \multicolumn{2}{c}{\textbf{7Scenes}}
  & \multicolumn{2}{c}{\textbf{Sintel}}
  & \multicolumn{2}{c}{\textbf{TUM}} \\
\cmidrule(lr){2-3}\cmidrule(lr){4-5}\cmidrule(lr){6-7}\cmidrule(lr){8-9}\cmidrule(lr){10-11}
  & ATE RMSE $\downarrow$ & AUC$_5$ $\uparrow$
  & ATE RMSE $\downarrow$ & AUC$_5$ $\uparrow$
  & ATE RMSE $\downarrow$ & AUC$_5$ $\uparrow$
  & ATE RMSE $\downarrow$ & AUC$_5$ $\uparrow$
  & ATE RMSE $\downarrow$ & AUC$_5$ $\uparrow$ \\
\midrule
            DUSt3R
                & 0.1127 & 10.86
                & 0.0285 & 55.80
                & 0.0379 & 21.33
                & 0.0851 & 10.62
                & 0.0207 & 7.00 \\
            VGGT
                & 0.0400 & 31.26
                & 0.0296 & 63.80
                & 0.0188 & 28.00
                & 0.0459 & 15.08
                & 0.0059 & 31.80 \\
            Pi3
                & 0.0464 & 34.20
                & 0.0152 & 71.80
                & 0.0183 & 28.33
                & 0.0283 & 19.38
                & 0.0036 & 31.80 \\
            Pow3r
                & 0.0882 & 15.13
                & 0.0313 & 56.40
                & 0.0262 & 19.00
                & 0.1093 & 8.31
                & 0.0146 & 9.40 \\
            VGGT-Omega
                & 0.0732 & 29.73
                & 0.0217 & 73.00
                & 0.0195 & 39.22
                & 0.0134 & 47.54
                & 0.0016 & 37.80 \\
            MapAnything
                & 0.0444 & 26.60
                & 0.0211 & 43.40
                & 0.0325 & 13.78
                & 0.0383 & 8.62
                & 0.0066 & 16.80 \\
\modelname & 0.0650 
& 4.40 
& 0.0522 & 8.60 & 0.0398 & 3.56 & 0.0588 & 0.77 & 0.0206 & 1.80 \\
\bottomrule
\end{tabular}
}
\caption{Comparison of model performance on multi-view pose estimation.}
\label{tab:pose}
\end{table}

\begin{table}[t]
\centering
\setlength{\tabcolsep}{3.5pt}
\renewcommand{\arraystretch}{1.12}
\resizebox{1\textwidth}{!}{%
\begin{tabular}{l cc cc cc cc cc}
\toprule
\multirow[t]{2}{*}{\textbf{Model}}
  & \multicolumn{2}{c}{\textbf{ScanNet++}}
  & \multicolumn{2}{c}{\textbf{Neural RGB-D}}
  & \multicolumn{2}{c}{\textbf{7Scenes}}
  & \multicolumn{2}{c}{\textbf{Sintel}}
  & \multicolumn{2}{c}{\textbf{TUM}} \\
\cmidrule(lr){2-3}\cmidrule(lr){4-5}\cmidrule(lr){6-7}\cmidrule(lr){8-9}\cmidrule(lr){10-11}
  & Abs Rel $\downarrow$ & Inlier Thresh $\uparrow$
  & Abs Rel $\downarrow$ & Inlier Thresh $\uparrow$
  & Abs Rel $\downarrow$ & Inlier Thresh $\uparrow$
  & Abs Rel $\downarrow$ & Inlier Thresh $\uparrow$
  & Abs Rel $\downarrow$ & Inlier Thresh $\uparrow$ \\
\midrule
            DUSt3R
                & 0.1727 & 0.2163
                & 0.3890 & 0.6217
                & 0.4244 & 0.4968
                & 2.0296 & 0.0700
                & 0.2408 & 0.1404 \\
            VGGT
                & 0.0782 & 0.3949
                & 0.3669 & 0.6489
                & 0.3979 & 0.4736
                & 1.4560 & 0.0872
                & 0.1274 & 0.2340 \\
            Pi3
                & 0.1011 & 0.3715
                & 0.3568 & 0.7910
                & 0.4779 & 0.5235
                & 0.5485 & 0.0820
                & 0.0829 & 0.2984 \\
            Pow3r
                & 0.1540 & 0.2648
                & 0.3805 & 0.5867
                & 0.4507 & 0.4504
                & 1.8443 & 0.0479
                & 0.2782 & 0.0933 \\
            VGGT-Omega
                & 0.0944 & 0.3743
                & 0.3594 & 0.8136
                & 0.4257 & 0.4417
                & 0.4948 & 0.2810
                & 0.0427 & 0.7202 \\
            MapAnything
                & 0.0848 & 0.3502
                & 0.3800 & 0.5670
                & 0.4676 & 0.3520
                & 0.9481 & 0.0945
                & 0.1597 & 0.1553 \\
\modelname & 0.1158 
& 0.2304 
& 0.4365 & 0.2690 & 0.4178 & 0.3435 & 0.8369 & 0.0618 & 0.2475 & 0.0727 \\
\bottomrule
\end{tabular}
}
\caption{Comparison of model performance on multi-view depth estimation.}
\label{tab:depth}
\end{table}

\begin{figure}[b]
  \centering
  \includegraphics[width=\textwidth]{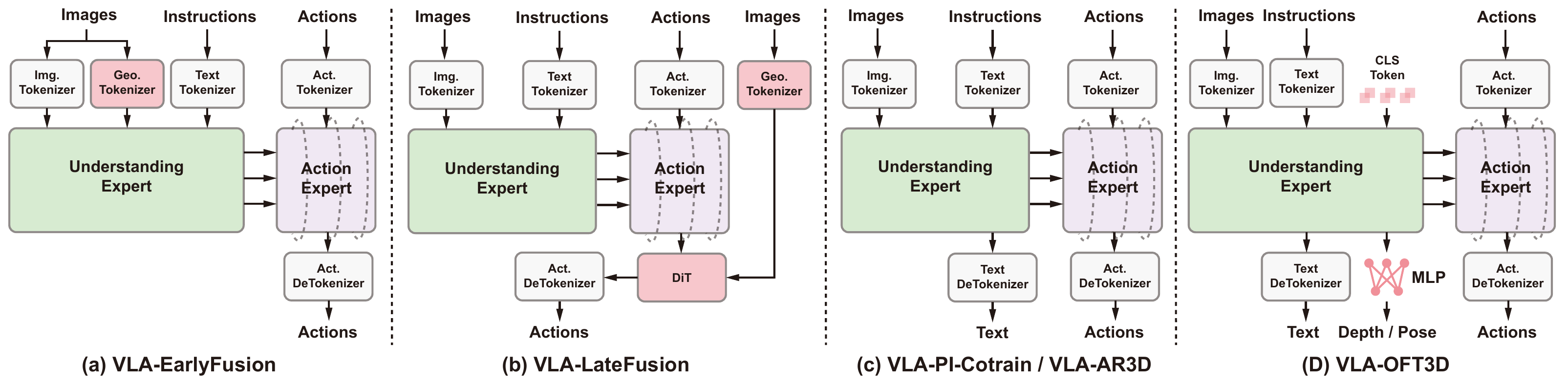}
  \caption{Illustration of VLA models.}
  \label{fig:vla}
\end{figure}

Under this challenging setting, \modelname recovers coherent scene geometry and achieves competitive camera translation accuracy, as measured by ATE RMSE. For dense depth prediction, \modelname also achieves comparable results across datasets. Qualitative depth estimation results are presented in Figure \ref{fig:depth}.
These results support the feasibility of learning multi-view geometric reconstruction within our model.
Nevertheless, pose AUC scores remain modest across benchmarks, indicating room for improvement in camera orientation estimation. One possible contributor is the attention formulation: \modelname processes frames using causal attention, whereas geometric foundation models use full attention, allowing unrestricted information exchange across views. We further fuse the predicted depth maps and camera poses to reconstruct multi-view point clouds, with qualitative results presented in Figure \ref{fig:point}.


\subsection{Action planning}
We compare against six VLA models, each described in detail below:
\begin{itemize}
    \item VLA-PI: Our baseline follows the $\pi_0$ formulation, using a dedicated action expert trained with a flow-matching objective to generate continuous actions \cite{black2410pi0}.
    \item VLA-PI-Cotrain: This variant extends VLA-PI by introducing additional visual question answering (VQA) supervision and jointly training on VQA and action prediction tasks, as shown in Figure \ref{fig:vla} (c).
    \item VLA-AR3D: Built upon VLA-PI-Cotrain, this variant incorporates a large amount of additional VQA data concerning multi-view reconstruction to strengthen geometric understanding through supervision in a vision–language format.
    \item VLA-OFT3D: Built upon VLA-PI-Cotrain, this variant introduces depth and pose prediction using the OFT \cite{community2026starvla, kim2025oft} formulation as additional geometric learning objectives, as shown in \ref{fig:vla} (d).
    \item VLA-EarlyFusion: This variant augments VLA-PI with an additional VGGT geometry encoder alongside the visual encoder, as shown in Figure \ref{fig:vla} (a). The extracted geometric features are added to the visual features, and the resulting fused representations are fed into the VLA model \cite{lin2025evo, abouzeid2025geoaware, zheng2026learning, yang2026understanding, yu2026_3dmix}.
    \item VLA-LateFusion: This variant extends VLA-PI with an additional diffusion transformer (DiT) \cite{peebles2023scalable} decoding module that operates on the features produced by the action expert, as shown in Figure \ref{fig:vla} (b). Conditioned on geometric features extracted by VGGT, this module further refines the action tokens \cite{zhang2025spatial, yang2026abot, yang2026understanding, yu2026_3dmix}.
\end{itemize}

\begin{figure}[t]
  \centering
  \includegraphics[width=0.9\textwidth]{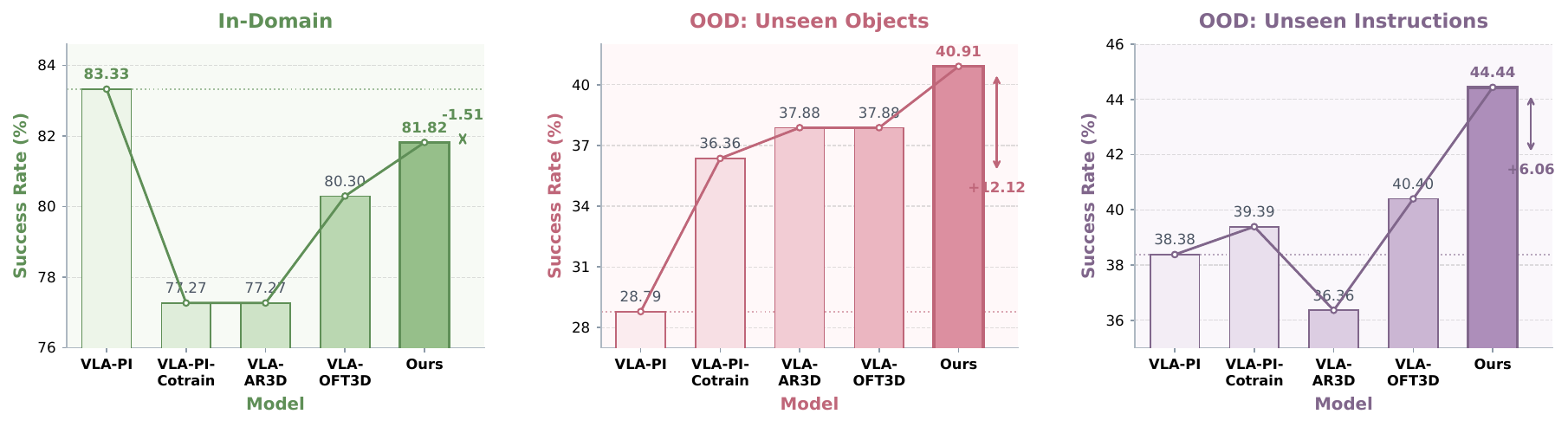}
  \caption{Success rate (SR) of VLA models.}
  \label{fig:sr}
\end{figure}

\begin{figure}[t]
  \centering
  \includegraphics[width=0.9\textwidth]{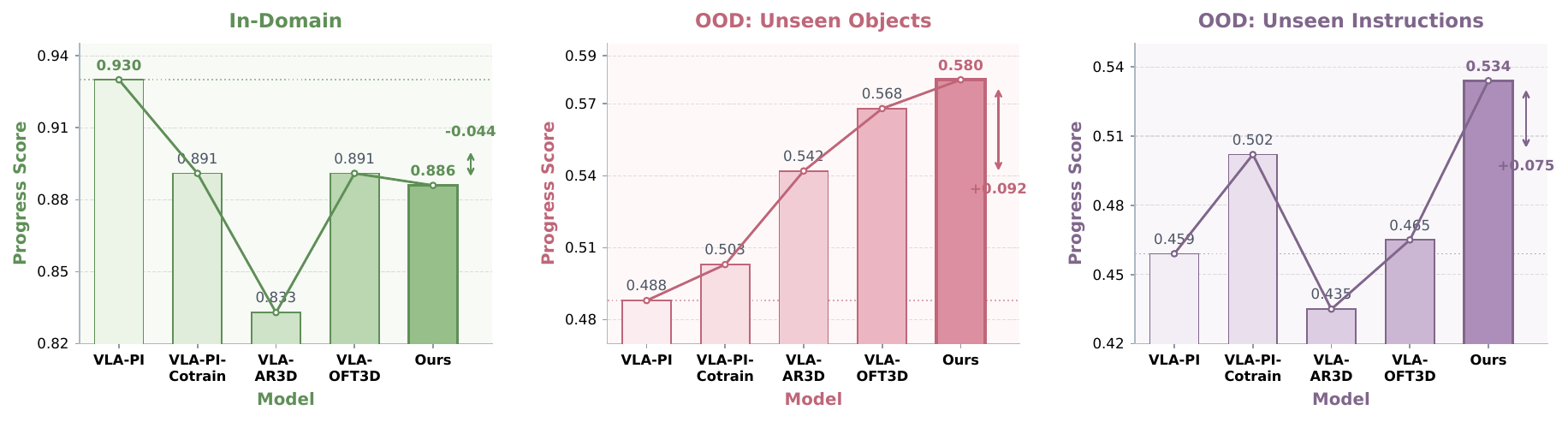}
  \caption{Progress score (PS) of VLA models.}
  \label{fig:ps}
\end{figure}

\begin{figure}[t]
  \centering
  \includegraphics[width=\textwidth]{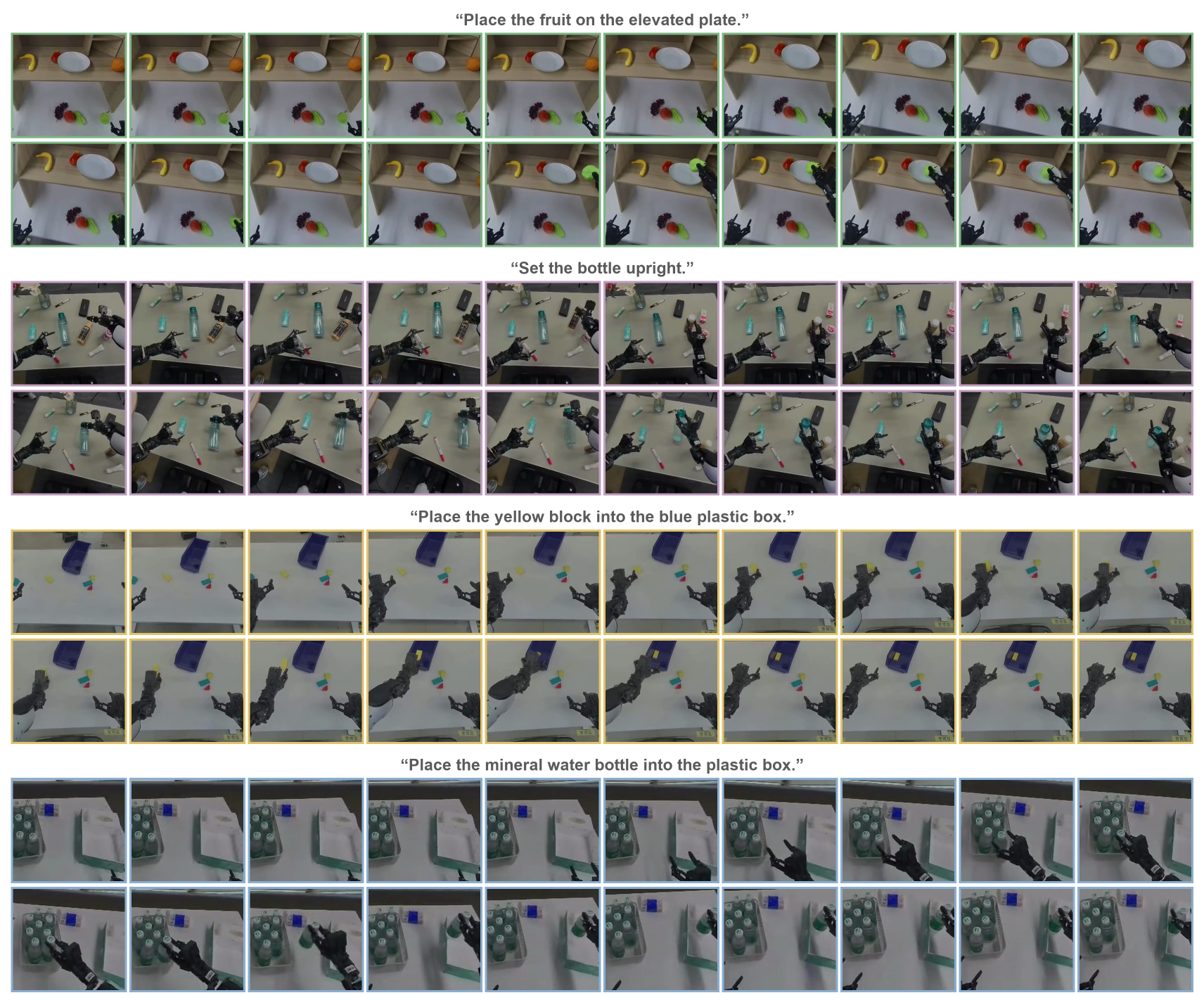}
  \caption{Visualization of robotic manipulation tasks executed by \modelname.}
  \label{fig:task}
\end{figure}

As shown in Figures \ref{fig:sr} and \ref{fig:ps}, the benefits of \modelname are most pronounced under distribution shifts. On unseen objects and unseen instructions, it achieves success rates of 40.91\% and 44.44\%, exceeding VLA-PI by 12.12 and 6.06 percentage points, respectively, with corresponding progress-score gains of 0.092 and 0.075. It also outperforms all other variants included in these Figures on both metrics in both out-of-domain (OOD) settings. A complementary task-level comparisons are provided in Tables \ref{tab:sr} and \ref{tab:ps}, where \modelname achieves the largest average improvements over VLA-PI in both success rate (+5.6 percentage points) and progress score (+0.046). However, VLA-AR3D provides no average success-rate improvement. Notably, despite achieving the strongest spatial question-answering results in Table \ref{tab:ab1}, VLA-AR3D does not deliver comparable manipulation gains, suggesting that proficiency in text-based spatial reasoning alone does not guarantee effective transfer to action. External geometric feature fusion also produces mixed results: VLA-LateFusion improves success rate by 2.8 percentage points, while VLA-EarlyFusion degrades both metrics. Although gains vary across tasks, these results support jointly learning dense geometry within the VLA backbone as an effective strategy for improving manipulation, particularly generalization to unfamiliar objects and instructions. The qualitative rollouts of \modelname across diverse real-world manipulation tasks are visualized in Figure \ref{fig:task}.

The descriptions of each robot manipulation task are provided below:
\begin{itemize}
    \item Plate stack: place the plate on the table / in the microwave / in the storage box. 
    \item Bottle upright: place the beverage bottle upright on the table.
    \item Desk trash: place the crumpled paper on the table into the trash bin.
    \item Get toast: remove the toast from the toaster.
    \item Cutlery box: place the cutlery in the storage box.
    \item Store blocks: place the building blocks in the storage box.
    \item Place blocks: take the building blocks out of the toy box and place them on the table.
    \item Cup coaster: place a cup in the center of the coaster.
    \item Refill water: place the bottle of mineral water in the storage box.
    \item Seasoning bottle: place the seasoning bottle on the spice rack.
    \item Fruit high plate: place the fruit on the table back into the elevated storage basket.
\end{itemize}

\subsection{Understanding}
We evaluate \modelname on a comprehensive suite of vision-language benchmarks, comparing it with leading embodied models (Cosmos3-Nano \cite{nvidia2026cosmos3}, RynnBrain \cite{dang2026rynnbrain}, Molmo2-ER \cite{fang2026molmoact2}, RoboBrain2.5 \cite{tan2026robobrain25}, and Qwen3.5 \cite{qwen2026qwen35}), flagship proprietary models (GPT-5.5 \cite{openai2026gpt55} and Gemini-3.1-Pro \cite{deepmind2026gemini31pro}), and specialized spatial reasoning models (SpaceLLaVA \cite{remyxai_spacellava}, SpatialRGPT \cite{cheng2024spatialrgpt}, DepthLM \cite{cai2026depthlm}, and $\mathrm{VLM}^{3}$ \cite{cai2026vlm3}).

\textbf{Spatial grounding.}
To assess whether learning dense 3D reconstruction improves 3D spatial understanding in vision–language tasks, we evaluate \modelname on depth estimation and camera pose estimation formulated as text-based spatial queries. These tasks require the model to express metric spatial information through language rather than produce dense geometric outputs. As shown in Table~\ref{tab:und}, the evaluation spans eight datasets covering indoor, outdoor, and synthetic scenes. Depth estimation is evaluated on Sintel \cite{butler2012naturalistic}, ScanNet++ \cite{yeshwanth2023scannet++}, NYUv2 \cite{silberman2012indoor}, KITTI \cite{geiger2012kitti}, and Bonn RGB-D \cite{palazzolo2019refusion}, while pose estimation is evaluated on TUM \cite{sturm2012benchmark}, Sintel \cite{butler2012naturalistic}, ScanNet++ \cite{yeshwanth2023scannet++}, 7Scenes \cite{shotton2013scene}, and Neural RGB-D \cite{azinovic2022neural}. \modelname achieves the highest average score on both tasks, outperforming the strongest baseline by $4.7$ points in depth estimation ($76.4$ vs. $71.7$) and $11.7$ points in pose estimation ($52.8$ vs. $41.1$). It ranks first on four of the five depth datasets and all five pose datasets, indicating consistent advantages across diverse scenes. These results suggest that the benefits of dense reconstruction learning extend beyond geometric output generation to 3D spatial understanding in vision–language tasks. 

\begin{table}[t]
\centering
\setlength{\tabcolsep}{3.5pt}
\renewcommand{\arraystretch}{1.12}
\resizebox{1.0\textwidth}{!}{%
\begin{tabular}{lcccccccccccc}
\toprule
\textbf{Model}
& \textbf{Plate stack}
& \textbf{Bottle upright}
& \textbf{Desk trash}
& \textbf{Get toast}
& \textbf{Cutlery box}
& \textbf{Store blocks}
& \textbf{Place blocks}
& \textbf{Cup coaster}
& \textbf{Refill water}
& \textbf{Seasoning bottle}
& \textbf{Fruit high plate}
& \textbf{Average} \\
\midrule
VLA-PI-Cotrain
                & +4.8 & -23.8 & -4.8 & +9.5
                & 0.0 & 0.0 & +4.8 & -14.3
                & -4.8 & +9.5 & +28.6 & +0.9 \\
VLA-AR3D
                & +14.3 & -4.8 & -23.8 & -9.5
                & -4.8 & -9.5 & -4.8 & -9.5
                & +4.8 & +14.3 & +33.3 & 0.0 \\
VLA-OFT3D
                & +14.3 & +4.8 & 0.0 & +4.8
                & 0.0 & -9.5 & +4.8 & 0.0
                & 0.0 & +4.8 & +4.8 & +2.6 \\
VLA-EarlyFusion
                & 0.0 & 0.0 & +27.3 & -4.0
                & -15.2 & +14.2 & -11.9 & -33.3
                & -18.2 & +18.2 & -51.5 & -6.8 \\
VLA-LateFusion
                & +14.3 & +10.0 & -9.5 & +19.0
                & +23.8 & -28.6 & -2.8 & +11.9
                & -11.9 & -7.1 & +11.9 & +2.8 \\
\modelname
                & +33.3 & 0.0 & -23.8 & -4.8
                & +9.5 & +9.5 & +9.5 & -9.5
                & 0.0 & +9.5 & +28.6 & \textbf{+5.6} \\
\bottomrule
\end{tabular}%
}
\caption{Success rate improvements of different VLA models over VLA-PI.}
\label{tab:sr}
\end{table}

\begin{table}[t]
\centering
\setlength{\tabcolsep}{3.5pt}
\renewcommand{\arraystretch}{1.12}
\resizebox{1\textwidth}{!}{%
\begin{tabular}{lcccccccccccc}
\toprule
\textbf{Model}
& \textbf{Plate stack}
& \textbf{Bottle upright}
& \textbf{Desk trash}
& \textbf{Get toast}
& \textbf{Cutlery box}
& \textbf{Store blocks}
& \textbf{Place blocks}
& \textbf{Cup coaster}
& \textbf{Refill water}
& \textbf{Seasoning bottle}
& \textbf{Fruit high plate}
& \textbf{Average} \\
\midrule
VLA-PI-Cotrain
                & -0.152 & -0.029 & -0.100 & +0.095
                & +0.019 & +0.038 & +0.029 & -0.057
                & -0.052 & +0.124 & +0.214 & +0.012 \\
VLA-AR3D
                & -0.019 & -0.048 & -0.271 & -0.067
                & +0.090 & -0.014 & -0.081 & -0.005
                & +0.038 & -0.105 & +0.238 & -0.022 \\
VLA-OFT3D
                & -0.067 & +0.057 & -0.067 & +0.048
                & +0.019 & -0.024 & +0.033 & 0.000
                & +0.010 & +0.095 & +0.052 & +0.014 \\
VLA-EarlyFusion
                & -0.036 & 0.000 & +0.164 & -0.060
                & -0.109 & -0.062 & -0.067 & -0.115
                & -0.258 & -0.030 & -0.285 & -0.078 \\
VLA-LateFusion
                & +0.074 & +0.253 & -0.121 & +0.036
                & +0.160 & -0.267 & -0.014 & +0.050
                & -0.205 & -0.050 & -0.057 & -0.013 \\
\modelname
                & +0.167 & -0.048 & -0.267 & -0.005
                & +0.095 & +0.067 & +0.062 & -0.019
                & 0.000 & +0.210 & +0.248 & \textbf{+0.046} \\
\bottomrule
\end{tabular}%
}
\caption{Progress score improvements of different VLA models over VLA-PI.}
\label{tab:ps}
\end{table}

\textbf{Robot-relevant Understanding \& Reasoning}. 
We further evaluate \modelname on understanding and reasoning tasks in embodied settings, using benchmarks including EgoTaskQA \cite{jia2022egotaskqa}, RoboSpatial-VQA \cite{song2025robospatial}, and Interactive Task Planning \cite{fang2025robix}.
To complement the Interactive Task Planning benchmark, we construct EQA (Embodied Question Answering) from real-world robot tele-operation data. EQA comprises four complementary diagnostic evaluations--Action Affordance, Manipulation Grounding, Task Verification, and Next Task Prediction--that respectively assess action preconditions, physical grounding, execution-outcome verification, and temporal task progression.
\modelname achieves an average score of 57.4 across the seven robot-relevant benchmarks, outperforming all other evaluated embodied models and exceeding the strongest embodied baseline, Qwen3.5, by 7.9 points. It also approaches the performance of flagship proprietary models, narrowing the gap to Gemini-3.1-Pro to 1.9 points. 
Overall, \modelname exhibits particular strengths in connecting scene understanding with manipulation context, assess task outcomes, and anticipate subsequent tasks.

\newcommand{\hdr}[1]{\begin{tabular}[c]{@{}c@{}}#1\end{tabular}}

\begin{table}[t]
\centering
\setlength{\tabcolsep}{2.8pt}
\renewcommand{\arraystretch}{1.02}
\resizebox{0.95\textwidth}{!}{%
\begin{tabular}{l *{8}{c}}
\toprule
\multirow[t]{2}{*}{\textbf{Benchmark}}
  & \multicolumn{6}{c}{\textbf{Embodied Models}}
  & \multicolumn{2}{c}{\textbf{Flagship Models}} \\
\cmidrule(lr){2-7}\cmidrule(lr){8-9}
  & \hdr{Cosmos3-Nano\\16B}
  & \hdr{RynnBrain\\30B-A3B}
  & \hdr{Molmo2-ER\\4B}
  & \hdr{RoboBrain2.5\\4B}
  & \hdr{Qwen3.5\\35B-A3B}
  & \hdr{{\modelname}\\{36B-A5B}}
  & \hdr{GPT-5.5}
  & \hdr{Gemini\\3.1-Pro} \\
\midrule
\multicolumn{9}{c}{\cellcolor{gray!15}\textbf{Depth Estimation}} \\
Sintel
  & 29.0 & 29.0 & 23.2 & 28.1 & 19.4 & 33.5 & {48.9} & 37.8 \\
ScanNet++
  & 61.1 & 63.6 & 61.4 & 54.2 & 59.6 & {89.3} & 76.8 & 72.8 \\
NYUv2
  & 60.9 & 63.5 & 64.5 & 60.6 & 53.6 & {86.3} & 76.6 & 74.0 \\
KITTI
  & 46.2 & 70.2 & 42.6 & 70.3 & 54.5 & {82.5} & 75.5 & 65.4 \\
Bonn
  & 69.0 & 81.2 & 73.5 & 70.7 & 56.8 & {90.5} & 80.8 & 77.9 \\
\textbf{Average}
  & 53.2 & 61.5 & 53.0 & 56.8 & 48.8
  & \textbf{76.4} & 71.7 & 65.6 \\
\midrule
\multicolumn{9}{c}{\cellcolor{gray!15}\textbf{Pose Estimation}} \\
TUM
  & 3.7 & 11.1 & 27.6 & 0.8 & 5.7 & {54.0} & 39.2 & 29.9  \\
Sintel
  & 6.2 & 15.2 & 48.4 & 3.5 & 15.1 & {51.8} & 47.0 & 36.4 \\
ScanNet++
  & 2.7 & 15.6 & 18.8 & 0.9 & 6.4 & {50.7} & 30.3 & 26.1 \\
7Scenes
  & 3.3 & 10.4 & 21.1 & 2.6 & 8.3 & {55.4} & 43.0 & 36.3 \\
Neural RGB-D
  & 4.2 & 10.8 & 12.1 & 5.6 & 18.5 & {51.9} & 46.1 & 40.4 \\
\textbf{Average}
  & 4.0 & 12.6 & 25.6 & 2.7 & 10.8
  & \textbf{52.8} & 41.1 & 33.8 \\
\midrule
\multicolumn{9}{c}{\cellcolor{gray!15}\textbf{Robot-relevant Understanding \& Reasoning}} \\
EgoTaskQA
  & 26.2 & {64.8} & 25.7 & 25.8 & 25.1 & 23.6 & 40.5 & 32.9 \\
RoboSpatial-VQA
  & 76.8 & 66.7 & 65.4 & 71.1 & 70.2 & 64.5 & {84.7} & 67.1 \\
EQA-Action Affordance
  & 59.2 & 43.3 & 49.2 & 60.0 & 68.3 & 65.8 & 70.0 & {71.7} \\
EQA-Manipulation Grounding
  & 14.4 & 19.7 & 19.7 & 16.7 & 28.8 & {73.5} & 45.5 & 37.9 \\
EQA-Task Verification
  & 11.8 & 32.3 & 48.0 & 53.8 & 55.1 & 75.9 & {76.1} & 71.9 \\
EQA-Next Task Prediction
  & 45.4 & 50.5 & 22.7 & 50.5 & 58.3 & 63.9 & 75.3 & {80.4} \\
EQA-Interactive Task Planning
  & 42.1 & 30.3 & 29.7 & 33.6 & 41.0 & 34.7 & 51.1 & {53.1} \\
\textbf{Average}
  & 39.4 & 43.9 & 37.2 & 44.5 & 49.5
  & 57.4 & \textbf{63.3} & 59.3 \\
\midrule
\multicolumn{9}{c}{\cellcolor{gray!15}\textbf{General Multimodal Understanding \& Reasoning}} \\
Core Cognition
  & 62.4 & 42.5 & 59.2 & 59.0 & 63.7 & 56.1 & 84.4 & {88.8} \\
MME
  & 86.4 & 84.8 & 68.4 & 81.1 & 82.8 & 75.1 & 89.1 & {91.9} \\
MMMU
  & 49.8 & 22.3 & 46.4 & 55.4 & 74.8 & 34.7 & 83.0 & {87.3} \\
MMStar
  & 64.5 & 68.1 & 59.9 & 57.3 & 72.0 & 52.9 & 81.6 & {84.4} \\
MMBench v1.1
  & 88.1 & 88.9 & 86.5 & 88.0 & 92.8 & 78.9 & 91.7 & {95.0} \\
CV-Bench
  & 75.5 & 67.4 & 87.4 & 63.8 & 87.4 & 62.3 & 88.1 & {89.5} \\
PixMo-Count
  & 63.0 & 45.9 & 85.8 & 58.1 & 73.6 & 50.7 & 84.6 & {90.1} \\
PointBench
  & 59.9 & 54.9 & 10.2 & {60.2} & 51.0 & 48.5 & 54.4 & 27.6 \\
Molmo2-Count
  & 21.5 & 22.5 & 17.7 & 22.7 & 15.6 & 21.0 & {40.0} & 27.4 \\
GST-Bench
  & 27.7 & 26.1 & 15.4 & 23.6 & 24.2 & 33.7 & {55.0} & 54.5 \\
EmbSpatial
  & 79.7 & 79.3 & 75.6 & 74.0 & 76.9 & 70.5 & {81.0} & 80.1 \\
\textbf{Average}
  & 61.7 & 54.8 & 55.7 & 58.5 & 65.0
  & 53.1 & \textbf{75.7} & 74.2 \\
\bottomrule
\end{tabular}%
}
\caption{Comparison of model performance across vision-language benchmarks.}
\label{tab:und}
\end{table}

\textbf{General Multimodal Understanding \& Reasoning}. 
Alongside its strong performance on geometric and robot-relevant tasks, \modelname maintains broad multimodal understanding and reasoning capabilities, achieving an average score of 53.1 across a suite of general-purpose benchmarks, including Core Cognition \cite{li2025corecognition}, CV-Bench \cite{tong2024cambrian1}, MME \cite{fu2023mme}, MMMU \cite{yue2024mmmu}, MMStar \cite{chen2024mmstar}, MMBench \cite{liu2024mmbench}, PixMo-Count \cite{deitke2025molmo}, PointBench \cite{cheng2025pointarena}, Molmo2-Count \cite{clark2026molmo2}, GST-Bench \cite{zhang2026gstbench}, and EmbSpatial \cite{du2024embspatial}.
This score is within 1.7 and 2.6 points of RynnBrain and Molmo2-ER, respectively, although it trails Qwen3.5 by 11.9 points. 
Notably, \modelname achieves the highest GST-Bench score among the evaluated embodied models (33.7).
These results show that strong 3D perception can coexist with broad multimodal competence, while leaving room to improve the balance between geometric specialization and general-purpose understanding.

\textbf{VLM$^3$ Depth Estimation Benchmark}. 
We further evaluate \modelname on the open-source spatial grounding benchmark introduced in VLM$^3$ \cite{cai2026vlm3}, which assesses the depth estimation capabilities of vision–language models. The evaluation covers diverse indoor and outdoor scenes from datasets including DDAD \cite{guizilini2020packnet}, nuScenes \cite{caesar2020nuscenes}, ETH3D~\cite{schops2017multi}, and ScanNet++~\cite{yeshwanth2023scannet++}, sunRGBD \cite{zhou2014learning}, iBims1 \cite{koch2018ibims}, and NYUv2 \cite{silberman2012indoor}. As shown in Table \ref{tab:vlm3}, \modelname achieves the highest average score among the evaluated embodied models, reaching 0.475 and surpassing the strongest competing baseline, RoboBrain2.5, by 0.089. 
It also substantially outperforms the evaluated spatial models, SpaceLLaVA and SpatialRGPT, which obtain average scores of 0.161 and 0.227, respectively. After fine-tuning on the corresponding training sets, \modelname further improves its average score to 0.916, exceeding both DepthLM (0.839) and VLM$^3$ (0.905). 
Together, these results demonstrate the effectiveness of \modelname for depth estimation and suggest that its geometric capabilities provide a strong foundation for further adaptation with task-specific supervision.

\begin{table}[t]
\centering
\setlength{\tabcolsep}{3.5pt}
\renewcommand{\arraystretch}{1.12}
\resizebox{0.8\textwidth}{!}{%
\begin{tabular}{lcccccccc}
\toprule
\textbf{Model}
  & \textbf{DDAD} & \textbf{NuScenes} & \textbf{ETH3D} & \textbf{ScanNet++} & \textbf{sunRGBD} & \textbf{iBims1} & \textbf{NYUv2} & \textbf{Average} \\
\midrule
\multicolumn{9}{c}{\cellcolor{gray!15}\textbf{Embodied Models}} \\
\text{Cosmos3-Nano-16B}
  & 0.216 & 0.187 & 0.294 & 0.403 & 0.389 & 0.324 & 0.336 & 0.307 \\
\text{RynnBrain-30B-A3B}
  & 0.287 & 0.379 & 0.450 & 0.437 & 0.339 & 0.541 & 0.365 & 0.400 \\
\text{Molmo2-ER-4B}
  & 0.184 & 0.174 & 0.232 & 0.344 & 0.355 & 0.341 & 0.359 & 0.284 \\
\text{RoboBrain2.5-4B}
  & 0.325 & 0.379 & 0.343 & 0.453 & 0.337 & 0.521 & 0.343 & 0.386 \\
\text{Qwen3.5-30B-A3B}
  & 0.237 & 0.270 & 0.343 & 0.410 & 0.366 & 0.417 & 0.411 & 0.351 \\
\text{\modelname-36B-A5B}
  & 0.222 & 0.157 & 0.366 & 0.521 & 0.591 & 0.791 & 0.676 & \textbf{0.475} \\
\midrule
\multicolumn{9}{c}{\cellcolor{gray!15}\textbf{Flagship Models}} \\
\text{Gemini-3.1-Pro}
  & 0.340 & 0.449 & 0.479 & 0.475 & 0.443 & 0.530 & 0.512 & 0.461 \\
\text{GPT-5.5}
  & 0.437 & 0.587 & 0.588 & 0.564 & 0.368 & 0.604 & 0.418 & \textbf{0.509} \\
\midrule
\multicolumn{9}{c}{\cellcolor{gray!15}\textbf{Spatial Models}} \\
\text{SpaceLLaVA-13B}
  & 0.067 & 0.083 & 0.090 & 0.269 & 0.233 & 0.208 & 0.178 & 0.161 \\
\text{SpatialRGPT-8B}
  & 0.046 & 0.100 & 0.220 & 0.346 & 0.369 & 0.240 & 0.265 & \textbf{0.227} \\
\midrule
\multicolumn{9}{c}{\cellcolor{gray!15}\textbf{Models Fine-tuned on Corresponding Training Set}} \\
\text{DepthLM-7B}
  & 0.747 & 0.865 & 0.718 & 0.850 & 0.859 & 0.920 & 0.915 & 0.839 \\
\text{VLM3-4B}
  & {0.818} & {0.970} & {0.810} & {0.976} & {0.867} & {0.960} & {0.935} & 0.905 \\
\text{\modelname-36B-A5B}
  & 0.684 & 0.980 & 0.953 & 0.919 & 0.908 & 0.998 & 0.970 & \textbf{0.916}  \\
\bottomrule
\end{tabular}%
}
\caption{Comparison of model performance on VLM$^3$ benchmarks.}
\label{tab:vlm3}
\end{table}

\subsection{Ablations}
We additionally conduct ablation studies on spatial grounding and general understanding benchmarks, comparing VLA-PI-Cotrain, VLA-AR3D, VLA-OFT3D, and \modelname, to investigate how different approaches to incorporating geometric learning affect spatial grounding and general vision–language performance.
On the spatial grounding benchmarks in Table \ref{tab:ab1}, VLA-AR3D achieves the highest average scores for both depth and pose estimation. This result is expected, given its use of the largest number of vision–language question–answer pairs concerning multi-view geometry, which closely align with the evaluation format. Compared with VLA-PI-Cotrain, \modelname improves the average depth estimation score from 75.4 to 76.4 and the average pose estimation score from 50.5 to 52.8. These gains suggest that learning dense geometric reconstruction supports spatial understanding expressed through vision–language tasks, providing evidence of transfer beyond the reconstruction objective itself. 
However, the results by VLA-OFT3D* suggest that effectively learning and expressing geometric information through the OFT formulation remains challenging under the evaluated setup.

The results in Table \ref{tab:ab2} reveal a modest trade-off between geometric specialization and general vision–language performance. VLA-PI-Cotrain achieves the highest average score of 55.7, whereas VLA-AR3D, VLA-OFT3D, and \modelname perform slightly worse. 
The modest declines across these variants suggest that general understanding capabilities are largely preserved when additional spatial and geometric learning objectives are introduced.


\begin{table}[t]
\centering
\setlength{\tabcolsep}{3.5pt}
\renewcommand{\arraystretch}{1.12}
\resizebox{1\textwidth}{!}{%
\begin{tabular}{l *{12}{c}}
\toprule
\multirow[t]{2}{*}{\textbf{Model}}
  & \multicolumn{6}{c}{\textbf{Depth Estimation}}
  & \multicolumn{6}{c}{\textbf{Pose Estimation}} \\
\cmidrule(lr){2-7}\cmidrule(lr){8-13}
  & \hdr{Sintel}
  & \hdr{ScanNet++}
  & \hdr{NYUv2}
  & \hdr{KITTI}
  & \hdr{Bonn}
  & \hdr{Average}
  & \hdr{TUM}
  & \hdr{Sintel}
  & \hdr{ScanNet++} 
  & \hdr{7Scenes} 
  & \hdr{Neural RGB-D} 
  & \hdr{Average}
  \\
\midrule
VLA-PI-Cotrain
  & 30.5 & 87.3 & 84.4 & 84.1 & 90.7 & 75.4 & 53.6 & 50.1 & 47.7 & 52.0 & 49.2 & 50.5 \\
VLA-AR3D
  & 35.8 & 88.7 & 87.0 & 88.4 & 92.1 & \textbf{78.4} & 57.0 & 55.2 & 54.5 & 62.5 & 51.0 & \textbf{56.0} \\
VLA-OFT3D
  & 26.7 & 84.9 & 86.1 & 82.4 & 84.7 & 73.0 & 52.4 & 51.3 & 48.6 & 53.2 & 50.3 & 51.2 \\
VLA-OFT3D*
  & 32.8 & 52.1 & 70.6 & 85.3 & 81.0 & 64.4 & 40.4 & 40.0 & 37.2 & 45.1 & 40.4 & 40.6 \\
\modelname
  & 33.5 & 89.3 & 86.3 & 82.5 & 90.5 & 76.4 & 54.0 & 51.8 & 50.7 & 55.4 & 51.9 & 52.8 \\
\bottomrule
\end{tabular}%
}
\caption{Comparison of model performance across spatial grounding benchmarks.
$^*$OFT evaluation.}
\label{tab:ab1}
\end{table}

\begin{table}[t]
\centering
\setlength{\tabcolsep}{3.5pt}
\renewcommand{\arraystretch}{1.12}
\resizebox{1\textwidth}{!}{%
\begin{tabular}{lcccccccccccc}
\toprule
\textbf{Model}
  & \textbf{Core Cognition} & \textbf{MME} & \textbf{MMMU} & \textbf{MMStar} & \textbf{MMBench v1.1} & \textbf{CV-Bench} & \textbf{PixMo-Count} & \textbf{PointBench} & \textbf{Molmo2-Count} & \textbf{GST-Bench} & \textbf{EmbSpatial} & \textbf{Average}\\
\midrule
\text{VLA-PI-Cotrain}
  & 59.5 & 75.7 & 40.0 & 59.2 & 81.8 & 70.9 & 48.6 & 51.8 & 17.3 & 34.8 & 72.7 & \textbf{55.7} \\
\text{VLA-AR3D}
  & 57.1 & 76.3 & 32.2 & 54.5 & 79.7 & 64.4 & 47.4 & 43.6 & 19.1 & 33.2 & 74.5 & 52.9 \\
\text{VLA-OFT3D}
  & 60.2 & 67.4 & 39.2 & 53.1 & 78.3 & 61.2 & 52.8 & 48.0 & 16.1 & 28.7 & 73.2 & 52.6 \\
\text{\modelname}
  & 56.1 & 71.6 & 34.7 & 52.9 & 78.9 & 62.3 & 50.7 & 48.5 & 21.0 & 33.7 & 70.5 & 52.8 \\
\bottomrule
\end{tabular}%
}
\caption{Comparison of model performance across understanding benchmarks.}
\label{tab:ab2}
\end{table}

\section{Related Work}


\noindent\textbf{VLA Models.}
Early VLA systems demonstrated that web-scale semantic knowledge can transfer to robotic control \cite{brohan2023rt}, while subsequent open and generative formulations scaled language-conditioned manipulation through autoregressive or continuous action decoding \cite{kim2024openvla,black2410pi0}. Recent models further advance embodied reasoning and planning \cite{fang2025robix}, architectural specialization and memory \cite{li2024vision,zhou2025vision,shi2026memoryvla,hu2026bagelvla}, and efficient or expressive action generation \cite{pertsch2025fast,wen2025dexvla}. Other efforts expand the sensory or predictive space with 3D, 4D, or motion representations \cite{zhen20243d,zhang20264d,wang2026lamp}. Nevertheless, most general-purpose VLAs still inherit representations optimized primarily for image--text alignment; stronger action heads or longer temporal context do not by themselves make metric scene structure observable to the policy. Semantic recognition can therefore remain disconnected from the depth, pose, and cross-view consistency required for precise manipulation. In contrast, our method treats geometry as a native learning target of a standard Mixture-of-Transformers VLA: lightweight geometric decoders expose dense metric structure from backbone states, while reconstruction, VQA, and action objectives are jointly optimized so that geometric perception is learned together with reasoning and control.

\noindent\textbf{Geometric Foundation Models.}
Geometric foundation models have transformed 3D perception from narrowly trained pipelines into transferable feed-forward predictors. DUSt3R casts multi-view reconstruction as dense pointmap regression \cite{wang2024dust3r}; VGGT unifies camera, depth, point, and tracking prediction within a global transformer \cite{wang2025vggt}; and Pi3 emphasizes permutation-equivariant multi-view geometry \cite{wang2026pi}. Follow-up systems incorporate camera and scene priors \cite{jang2025pow3r}, improve robustness and scale \cite{wang2026vggt}, or target general metric reconstruction from arbitrary views \cite{lin2025depth,keetha2026mapanything}. Together, these models provide strong geometric correspondences and dense scene estimates across diverse imagery, making them attractive spatial priors for downstream agents. Their specialization, however, is also a boundary: they are trained primarily to recover geometric outputs rather than to ground language, reason about robot tasks, or generate actions, and their typically bidirectional multi-view processing does not directly match causal embodied inference. Placing such a model beside a VLA therefore leaves unresolved how geometric features should align with semantic and action representations. Our work instead internalizes reconstruction within the embodied backbone. A lightweight DPT decoder and token-level pose and scale heads supervise metric depth and camera motion directly from VLA hidden states, allowing a single causal MoT model to acquire native geometric competence without carrying a separate geometric foundation model at inference time.

\noindent\textbf{3D Fusion for VLM/VLA Models.}
Research connecting 3D perception to embodied models differs not only in representation, but also in where geometry enters the learning system. Some methods supply depth, stereo, point-cloud, or scene representations directly to language or action models \cite{hong20233d,fu2024scene,zheng2025video,han2026stereopolicy,li2026pointvla,deng2025stereovla}. Others import features from pretrained geometric networks, with fusion occurring before the embodied backbone, near the action decoder, or at multiple stages \cite{lin2025evo,abouzeid2025geoaware,zheng2026learning,zhang2025spatial,yang2026abot,yang2026understanding,yu2026_3dmix}. These designs can improve spatial sensitivity, yet they couple the policy to additional encoders, adapters, and feature-alignment choices, increasing system complexity and leaving geometric knowledge auxiliary to the VLA itself. A complementary direction converts metric or spatial supervision into language-based question--answer pairs \cite{cai2026depthlm,cai2026vlm3}. Such supervision is scalable and compatible with standard VLM interfaces, but success at emitting textual 3D answers need not produce dense, action-usable scene representations. Our formulation bridges this divide without treating geometry as an additional input modality, an external expert feature, or a purely linguistic target. Starting from a standard MoT VLA, we append only lightweight geometric decoders and jointly optimize dense multi-view reconstruction, VQA, and continuous action generation. Geometry thus becomes an intrinsic capability of the embodied backbone, establishing a direct learning path from metric spatial structure to reasoning and physical control.

\section{Conclusion}
We presented \modelname, a multi-modal VLA model that acquires geometrically grounded spatial intelligence without relying on an external geometric foundation model or invasive architectural modifications. By jointly optimizing vision-language understanding, dense multi-view reconstruction, and continuous action generation, \modelname learns to reconstruct dense metric 3D geometry, reason about spatial relationships, and transfer these capabilities to real-world manipulation. We hope \modelname provides a scalable foundation for developing VLA models that perceive, reason about, and act within the 3D physical world through a unified architecture.
Future work will scale training to more diverse platforms and environments, improve inference efficiency, extend geometric modeling toward temporally consistent dynamic scenes, and incorporate stronger memory and planning mechanisms for long-horizon, cross-embodiment manipulation.


\section{Contributions and Acknowledgements}
\label{sec:author}

\textbf{Authors}
Ziyang Liu\textsuperscript{1,\(\dagger\)},
Jamie Tang\textsuperscript{2, \(\dagger\)},
Yongyao Jiang\textsuperscript{1,\(\dagger\)},
Tian Lan\textsuperscript{1,\(\dagger\)},
Hongbin Xu\textsuperscript{1,*},
Yanzhao Yu\textsuperscript{2},
Sheng Chen\textsuperscript{1,\(\ddagger\)},
Peiyu He\textsuperscript{1},
Xue Xu\textsuperscript{1},
Yuanfei Luo\textsuperscript{1},
Xueqian Wang\textsuperscript{2},
Yangang Zhang\textsuperscript{1},
and Hang Li\textsuperscript{1}.

\textbf{Affiliations}
\textsuperscript{1}ByteDance Seed,
\textsuperscript{2}Tsinghua University.

\textbf{Acknowledgements}
We thank Wei Li and Mingyao Li for providing robot-relevant understanding data. We thank Yufei Liu, Junhui Zhang, and Zibo Chen for their help with real-world robot inference and deployment. We thank Yansheng Wang and Liuchun Yuan for their help with action policy training. We thank Jinhua Wangfang for his valuable suggestions on this work.

\vspace{0.4em}
\(\dagger\) Equal contribution, \(\ddagger\) Project lead, * Corresponding author.

\clearpage

\bibliographystyle{unsrtnat}
\bibliography{main}

\clearpage


\end{document}